\documentclass{article}

\usepackage[preprint]{corl_2026} % Uncomment for pre-prints (e.g., arxiv); This is like ``final'', but will remove the CORL footnote.
\usepackage{graphicx}
\usepackage{float}
\usepackage{xcolor}
\usepackage{booktabs}
\usepackage{graphicx}
\usepackage{caption}
\usepackage{threeparttable}
\usepackage{wrapfig}
\usepackage{amsmath}
\usepackage{hyperref}
\usepackage{enumitem}

\definecolor{ruogu_color}{RGB}{84,130,53}

\definecolor{zhenyu_color}{RGB}{102,192,255}

\definecolor{citecolor}{HTML}{0071BC}
\hypersetup{colorlinks,linkcolor={red},citecolor={citecolor}}

\title{Handroid: Bridging Dexterous Hand and Humanoid}

\author{
Ruogu Li$^{1}$, Chenyang Ma$^{1}$, Sikai Li$^{1}$, Zhenyu Wei$^{1}$, Yunchao Yao$^{1}$, \\ 
\textbf{Haochen Shi$^{2}$, C. Karen Liu$^{2}$, Shuran Song$^{2}$, Mingyu Ding$^{1}$} \\
$^1$University of North Carolina at Chapel Hill \quad $^2$Stanford University
\\[0.5em]
Project webpage: \url{https://handroid.org}
}

\begin{document}
\maketitle
\vspace{-1.2em}

{
\centering
\vspace{-10pt}
\includegraphics[width=0.96\linewidth]{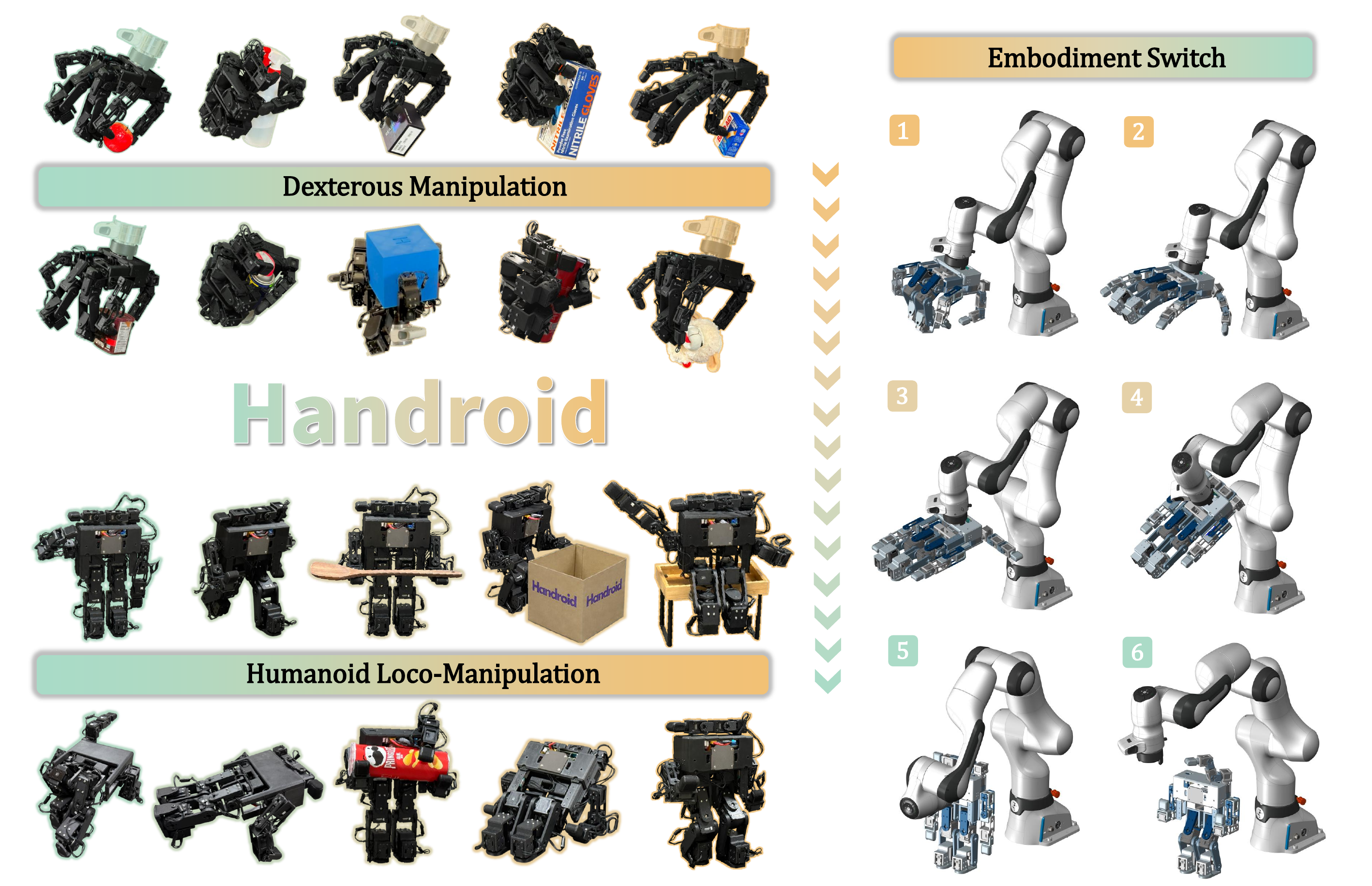}
\captionsetup{skip=3pt}
\captionof{figure}{\textbf{Left:} Handroid demonstrates dexterous manipulation in the hand embodiment and locomotion and loco-manipulation in the humanoid embodiment.
\textbf{Right:} Embodiment switching between the dexterous hand and humanoid morphologies.}
\label{fig:teaser}
\par
}

\begin{abstract}
Dexterous hands and humanoid robots are typically developed as distinct embodiments: the former enable contact-rich manipulation at the object scale, whereas the latter provide mobility and whole-body interaction in human-centered environments. We introduce \textbf{Handroid}, a desktop-scale dual-embodiment robot that integrates both capabilities within a single reconfigurable platform. Handroid reuses one 27-DoF electromechanical body as either a dexterous hand or a desktop humanoid, measuring 0.33 m in height and 2.05 kg in weight. In the dexterous hand embodiment, 20 DoFs form an anthropomorphic hand closely matching the kinematic structure of the human hand. In the humanoid embodiment, the same articulated modules are reconfigured into a humanoid with a head, arms, and legs, including a 12-DoF lower-limb structure for locomotion and whole-body motion. Handroid further provides a unified control and learning framework supporting hand teleoperation, dexterous grasping, in-hand manipulation, humanoid locomotion, gait generation, and interactive motion authoring. We validate the platform through real-world dexterous manipulation, reinforcement-learning-based locomotion, keyframe motion deployment, and a long-horizon task involving embodiment reconfiguration, locomotion, docking, and dexterous pick-and-place. These results position Handroid as a compact and reproducible platform for advancing morphology-reconfigurable robotics and cross-embodiment robot learning. 
% Project webpage: \url{https://handroid.org}.
\end{abstract}

% Two or three meaningful keywords should be ad{ded here

%===============================================================================
\section{Introduction}
\vspace{-10pt}

Morphology fundamentally defines a robot's functional boundaries: what it can reach, how it moves, what it can sense, and how it physically interacts with the environment~\cite{west1997general,biewener2005biomechanical}. Dexterous hands and humanoid robots represent two complementary forms of embodied capability. Dexterous hands provide compact, high-DoF structures for fine-grained, contact-rich manipulation, supporting grasping, in-hand reorientation, tool use, and delicate interaction~\cite{wei2024dro,napier1956prehensile,landsmeer1962power,marzke1997precision,santello1998postural,wei2026one}. Humanoid robots, by contrast, provide whole-body mobility and body-scale interaction in human-centered environments, enabling behaviors such as walking, squatting, reaching, and carrying~\cite{peng2018deepmimic,peng2021amp,he2024learning,he2024omnih2o,yang2025omniretarget,he2025asap}. The former emphasizes local dexterity, whereas the latter enables global mobility.

Despite this complementarity, the two morphologies are typically developed as independent platforms. Dexterous hands are commonly mounted on fixed-base robot arms, providing precise manipulation within a limited workspace. Humanoid robots can navigate and interact throughout an environment, but their hands are often underactuated or insufficiently dexterous for fine manipulation. Consequently, mobility and dexterity remain largely separated across robotic systems.

This separation raises a broader question for robot design and learning: \textit{can morphology be reused across embodiments rather than treated as a fixed property of a single robot?} Although hands and humanoid bodies differ in appearance and function, both can be viewed as collections of articulated kinematic chains connected to a central body. Fingers, arms, and legs differ in scale and task role, yet share common structural elements, including joint arrangements, contact geometry, actuation, sensing, and coordinated control. Reusing these elements across embodiments could enable a single physical system to support both contact-rich manipulation and whole-body mobility, while providing a common platform for studying control and learning across morphologies.

Motivated by this perspective, we introduce \textbf{Handroid}, a desktop-scale dual-embodiment robot that realizes morphology reuse in hardware. A single 27-DoF electromechanical body can be reconfigured as either a dexterous hand or a desktop humanoid. Handroid is 0.33~m tall and weighs 2.05~kg. In the Dexterous-Hand embodiment, 20 actuated joints form an anthropomorphic hand structure that closely follows the kinematic configuration of the human hand, supporting coordinated finger motion and manipulation tasks including pouring, tissue pulling, cup stacking, grasping, and in-hand cube reorientation. In the Humanoid embodiment, the same articulated modules are repurposed as the head, arms, and legs, including a 12-DoF lower-limb structure that supports walking, turning, squatting, stepping, and postural transitions. A pair of compact sliding mechanisms enables physical reconfiguration between the two embodiments.

Handroid further provides a unified control and learning stack across both embodiments. The system supports VR-based arm-hand teleoperation, object-conditioned dexterous grasping, reinforcement learning for in-hand reorientation, humanoid locomotion, ZMP-guided gait generation, and Viser-based motion authoring. These components share common sensing, control, simulation, and deployment interfaces, enabling demonstration collection, policy training, motion generation, real-world execution, and embodiment switching on the same platform.

We evaluate Handroid through dexterous manipulation and humanoid motion. In the Dexterous-Hand embodiment, we demonstrate fine-grained teleoperation, object-conditioned grasping learned from 100 demonstrations, and real-world in-hand cube reorientation. In the Humanoid embodiment, we evaluate ZMP-guided reinforcement learning, reference-free velocity control, and keyframe-authored motions, including walking, turning, squatting, pull-ups, push-ups, and pick-and-place. Finally, we demonstrate a long-horizon task combining embodiment switching, humanoid locomotion and object interaction, re-docking with a Franka robot arm, and dexterous pick-and-place. Together, these experiments establish Handroid as a compact and reproducible platform for studying morphology reuse, cross-embodiment learning, and integrated mobility and manipulation.

In summary, the main contributions of this work are threefold:
\begin{itemize}[leftmargin=*, itemsep=0.3em, topsep=0.3em]
    \item We design and open-source Handroid, a 27-DoF desktop-scale robot that reconfigures between Dexterous-Hand and Humanoid embodiments using a shared electromechanical system.
    \item We develop a unified learning and control stack covering teleoperation, imitation learning, reinforcement learning, gait generation, and interactive motion authoring across both embodiments.
    \item We validate Handroid through real-world dexterous manipulation, humanoid locomotion, motion deployment, and long-horizon cross-embodiment execution, demonstrating its potential as a compact and reproducible platform for studying morphology reuse, robot learning, and loco-manipulation.
\end{itemize}
\section{Related Work}

\noindent \textbf{Dexterous Hands.}
Dexterous hands are high-DoF end effectors designed for contact-rich manipulation and robot learning. Existing systems can be broadly categorized by their actuation mechanisms into tendon-driven and motor-driven designs. Tendon-driven platforms, such as Shadow Hand~\cite{shadowhand}, the ADROIT research hand~\cite{Kumar2014RealtimeBS}, and Xynova Flex 1~\cite{xynovaflex1}, achieve anthropomorphic structures with many actuated joints, but often suffer from friction, hysteresis, cable elasticity, and calibration or maintenance challenges that affect accuracy and repeatability. Motor-driven hands, including Wuji Hand 2~\cite{wujihand2} and SharpaWave~\cite{sharpawave}, provide more direct transmission and precise control, but their high cost limits accessibility. Among open-source platforms, LEAP Hand~\cite{shaw2023leaphand} is widely adopted for its low cost and mature learning ecosystem, although its morphology differs substantially from the human hand. Other systems, including D'Manus~\cite{bhirangi2024feelsdexteroushandlargearea}, ISyHand~\cite{richardson2025isyhanddexterousmultifingerrobot}, and ORCA~\cite{christoph2025orca}, explore different trade-offs in actuation, sensing, robustness, and structure. However, jointly achieving high DoF, precise control, and low cost remains challenging. Unlike these single-purpose end effectors, Handroid realizes the dexterous hand as one embodiment of a shared reconfigurable body that can also form a humanoid.

\noindent \textbf{Humanoids.}
Humanoid robots exhibit diverse morphologies and capabilities depending on their physical scale, structural design, and target applications. Existing platforms can be broadly grouped into full-size, half-size, and mini-size systems. Full-size humanoids, including Unitree H1~\cite{unitree}, Unitree H2~\cite{unitreeh2}, Unitree G1~\cite{unitreea}, Figure~\cite{figure}, Boston Dynamics Atlas~\cite{atlas}, Tesla Optimus~\cite{ai}, Digit~\cite{agility}, and Fourier GR-1~\cite{fourierrobotics}, mainly target locomotion, whole-body control, and deployment in human-scale environments. Half-size systems, such as Booster T1~\cite{humanoid}, Berkeley Humanoid~\cite{liao2024berkeleyhumanoidresearchplatform}, Berkeley Humanoid Lite~\cite{chi2025demonstratingberkeleyhumanoidlite}, iCub~\cite{icub}, Duke Humanoid~\cite{xia2025dukehumanoiddesigncontrol}, and BRUCE~\cite{9811790}, reduce cost, risk, and operational complexity while retaining bipedal locomotion and learning-based control capabilities. Mini-size platforms, including NAO H25~\cite{nao}, ROBOTIS OP3~\cite{nameintroduction}, K-Scale Zeroth~\cite{zeroth}, and ToddlerBot~\cite{shi2025toddlerbot}, further lower the barrier to experimentation and support education, rapid algorithm validation, and reproducible research. In contrast to these fixed-morphology systems, Handroid's humanoid embodiment reuses the same articulated modules as its dexterous hand, combining a safe desktop scale with a high-DoF structure for locomotion and whole-body control.

\noindent \textbf{Learning Methods.}
Beyond hardware, both dexterous manipulation and humanoid locomotion have increasingly relied on data and simulation-driven control. In dexterous manipulation~\cite{yao2026dexverse}, one line of work learns hand-object skills from teleoperation and retargeting~\cite{handa2020dexpilot,qin2023anyteleop,xu2025dexumi}, mocap demonstrations~\cite{wang2024dexcap,zhao2025dexh2r}, or object-conditioned datasets and policies~\cite{chi2025diffusion,wang2023dexgraspnet,li2023gendexgrasp,xu2023unidexgrasp,debakker2025scaffolding,zhong2025grasp2grasp,wei2024dro,huang2026dexcompose,liang2025dexhanddiff}. Another line focuses on in-hand manipulation, where policies are trained with RL and object-centric rewards~\cite{rajeswaran2018learning,openai2019rubiks,openai2020learning, huang2026dexcompose}, rapid adaptation or multimodal sensing~\cite{qi2023hora,qi2023rotateit, ma2026current}, and sim-to-real randomization~\cite{handa2023dextreme,chen2023visual,lin2025simtoreal}. For humanoid locomotion, learning methods similarly follow two directions: reference-based approaches imitate motion references~\cite{peng2018deepmimic,peng2021amp} or retarget human demonstrations into robot-compatible trajectories~\cite{he2024learning,he2024omnih2o,yang2025omniretarget}, while reference-free and real-world adaptation approaches learn locomotion and loco-manipulation from command rewards, privileged critics, curricula, domain randomization, force interaction, and sim-to-real adaptation~\cite{gu2024humanoidgym,he2025asap,zhang2025falcon,hu2025robot, li2026coordex, li2026anybodyfreeformwholebodyhumanoid}. 
Together, these methods provide effective tools for dexterous manipulation and locomotion, but they are typically designed around a single fixed embodiment. In contrast, Handroid provides a shared, reconfigurable platform on which both manipulation and locomotion learning can be studied across different robot configurations.
\section{Design}
\subsection{Structure Design}

\begin{figure}[t]
    \centering
    \includegraphics[width=\textwidth]{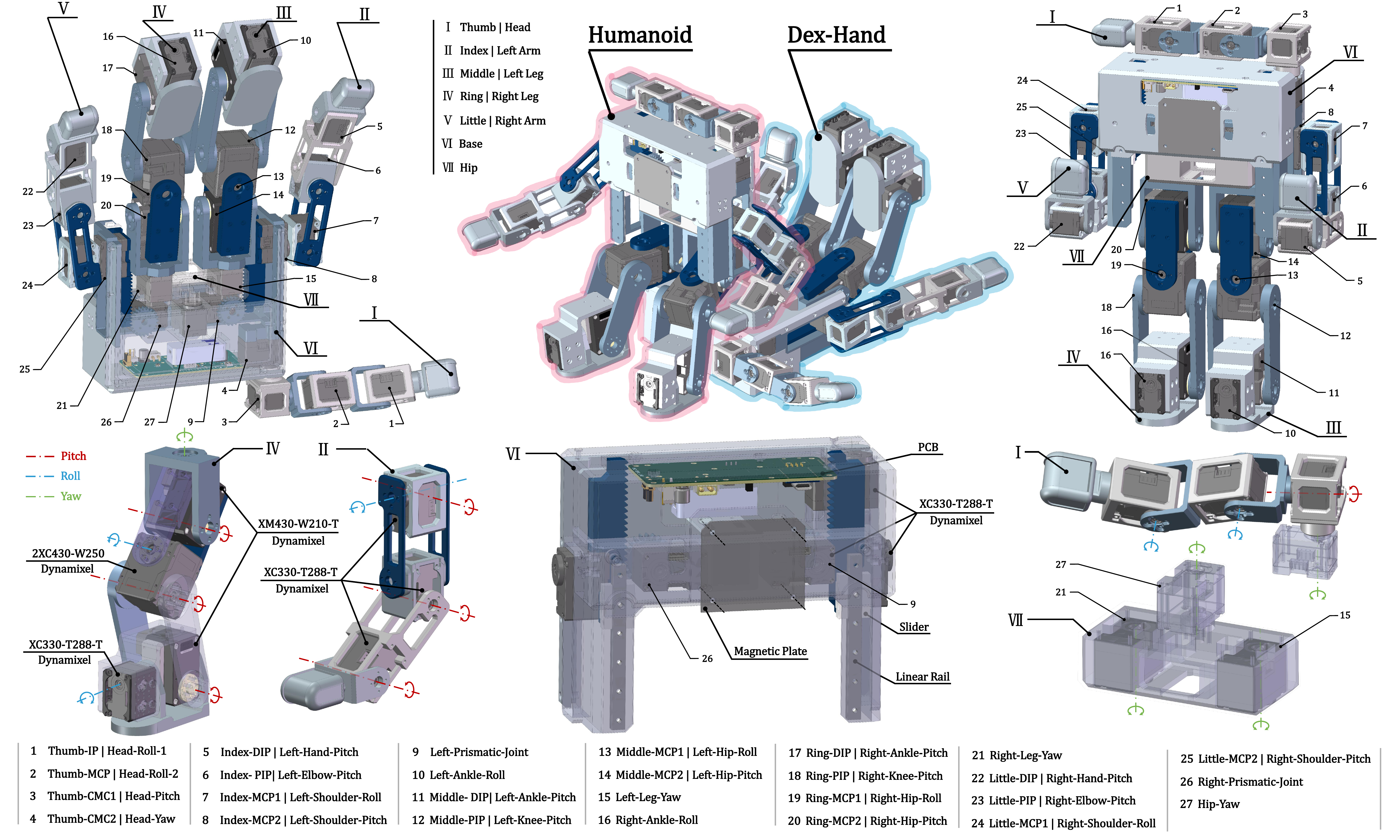}
    \vspace{-8pt}
    \caption{\textbf{Handroid Structure Design.} Modules I--V correspond to the five fingers in the hand embodiment and to the head, left arm, left leg, right leg, and right arm in the humanoid embodiment, while modules VI and VII form the base and hip. We show the module mapping, joint numbering, actuator layout, and pitch/roll/yaw axes. Joints 9 and 26 drive the prismatic sliding mechanism for embodiment switching.}
    \vspace{-12pt}
    \label{fig:Structure Design}
\end{figure}

From a morphological perspective~\cite{li2026sldprtnetlargescalemultimodaldataset}, the human body and hand can both be abstracted as branching topologies in which multiple articulated chains extend from a compact central structure. In particular, the body's torso-to-limb organization is analogous to the hand's palm-to-finger organization. This analogy motivates Handroid's central design principle: the same articulated modules can assume different functional roles across embodiments.

Guided by this principle, we design Handroid as a desktop-scale dual-embodiment robot with 27 actuated DoFs, as illustrated in Fig.~\ref{fig:Structure Design}. In the Dexterous-Hand embodiment, 20 articulated DoFs are distributed across five digits, with each digit providing one abduction--adduction DoF and three flexion--extension DoFs. This anthropomorphic kinematic arrangement approximates a commonly used 21-DoF human-hand model and supports thumb opposition, multi-finger coordination, grasping, and in-hand manipulation. In the Humanoid embodiment, 25 articulated DoFs are distributed across a 4-DoF head (Module~I), two 4-DoF arms (Modules~II and~V), two 6-DoF legs (Modules~III and~IV), and a 1-DoF central hip joint (Module~VII), while Module~VI serves as the central torso. The remaining two actuated DoFs are prismatic joints dedicated to embodiment reconfiguration. Together, the two legs provide 12 lower-limb DoFs that approximate the principal joint motions of the human lower limbs. The corresponding hand and
body range-of-motion (RoM) comparisons are respectively shown in
Figs.~\ref{fig:range-motion-hand} and~\ref{fig:range-motion-humanoid}.

The two compact linear reconfiguration mechanisms associated with joints~9 and~26 are integrated within Module~VI. Each mechanism uses a rack-and-pinion transmission to convert actuator rotation into translation along a rigid linear guide rail. During the transition from the Humanoid embodiment to the Dexterous-Hand embodiment, these mechanisms translate Modules~II and~V downward to their hand-configuration positions, where they serve as the index- and little-finger modules, respectively. The embodiment transition therefore requires only controlled module repositioning, without hardware replacement.

\begin{wrapfigure}{r}{0.5\linewidth}
    \centering
    \vspace{-14pt}
    \includegraphics[width=\linewidth]{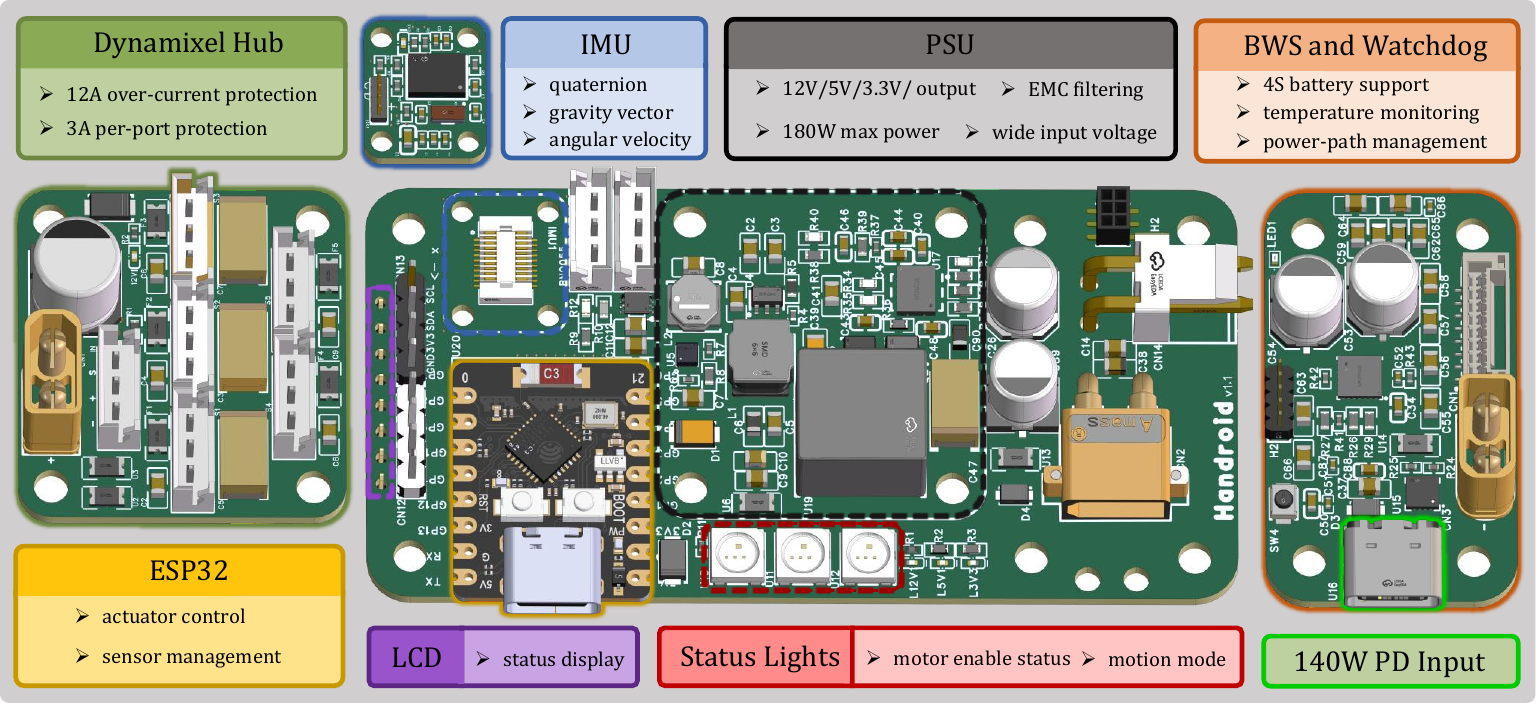}
    \vspace{-8pt}
    \caption{\textbf{Electrical Design.} A compact mainboard integrates control, sensing, power management, and status monitoring for Handroid's dual-embodiment operation.}
    \vspace{-10pt}
    \label{fig:pcb_design}
\end{wrapfigure}

\subsection{Electrical Design}

Handroid’s electrical system is designed as a compact integrated backbone for the dual-embodiment platform, unifying actuation control, power delivery, wireless communication, onboard sensing, and hardware-state monitoring within the limited internal volume of the robot (see Fig. \ref{fig:pcb_design}). A custom mainboard coordinates all Dynamixel actuators through a TTL bus, streams robot states to a host computer over Wi-Fi, and supports remote operation and onboard execution of basic motion primitives. The system is designed for both tethered development and untethered operation, with onboard power and temperature monitoring to protect the hardware during long experiments. Together with distributed inertial sensing and actuator feedback, this electrical architecture provides the proprioceptive and safety-critical interface required for dexterous manipulation, loco-manipulation, and embodiment-switching tasks on the same 27-DoF electromechanical system.

\noindent \textbf{Stacked mainboard architecture.} The Handroid mainboard adopts a vertically stacked modular architecture to increase usable integration area without enlarging the footprint of the robot. Functional modules are mounted above the base board using copper standoffs, which provide both mechanical support and power connections, while the elevated layout improves thermal dissipation around power-related components. The mainboard footprint is 40 mm × 80 mm and integrates the core electrical functions required by the 27-DoF system.

\noindent \textbf{Power management.} Handroid supports both battery-powered and externally powered operation. It can operate untethered from onboard batteries or be powered by a standard Power Delivery charger with input power up to 140 W. The same PD input can charge the onboard batteries, which simplifies long experiments and development without removing the batteries. An onboard STM32 monitors the power supply unit and battery temperature in real time, allowing the system to regulate operating load and reduce the risk of excessive power draw or overheating.

\noindent \textbf{Control and sensing.} For control and communication, the mainboard integrates an ESP32-S3 controller that commands all Dynamixel actuators through a TTL bus and reads back motor states, including joint position, velocity, load, and temperature. The controller stores basic motion primitives for standalone execution without a host computer, supports gamepad-based remote operation, and streams robot-state data to a host computer over Wi-Fi for higher-level policy deployment. For onboard sensing, Handroid uses compact IMU modules. One IMU is mounted on the mainboard to provide body orientation, gravity direction, and angular velocity, while additional IMUs are mounted at the index and middle fingertips, which correspond to the feet in the Humanoid embodiment. These sensing and feedback channels provide the proprioceptive interface used by both manipulation and locomotion controllers.

\section{Control}

Handroid exposes a single control and learning stack shared across both embodiments. The same actuators, proprioceptive channels, and communication interface serve two families of controllers: the Dexterous-Hand embodiment targets contact-rich manipulation, while the Humanoid embodiment targets locomotion and whole-body manipulation. Reusing one hardware and sensing backbone lets a controller developed for one morphology inherit the same interfaces in the other, which is what makes long-horizon tasks that span both embodiments implementable on a single robot. Detailed implementations are provided in Appendix~\ref{app:hand-control} and~\ref{app:humanoid-control}.
\subsection{Dexterous Hand}
\noindent\textbf{Teleoperation.}
For the Dexterous Hand embodiment, we build an Apple Vision Pro-based teleoperation interface for collecting coordinated arm-hand demonstrations. The interface retargets the operator's hand motion to Handroid and maps wrist motion to the end-effector command of a Franka Research 3 arm, allowing the external arm and Handroid to be controlled as a unified manipulation system~\cite{handa2020dexpilot,qin2023anyteleop}. This setup enables real-time demonstration collection for grasping, placing, and object interaction tasks.

\noindent \textbf{Dexterous grasping.} From teleoperated demonstrations, we train an object-conditioned diffusion policy for dexterous grasping~\cite{chi2025diffusion}. The policy conditions on object geometry and recent robot proprioception to generate arm-hand action chunks, enabling Handroid to grasp objects with varied geometries and poses~\cite{wang2023dexgraspnet,li2023gendexgrasp,xu2023unidexgrasp}.

\noindent \textbf{In-hand reorientation.}
Beyond grasping, we evaluate Handroid's in-hand manipulation capability through cube reorientation. We train an RL policy in simulation and deploy it to the real robot~\cite{openai2020learning,qi2023hora,handa2023dextreme,chen2023visual}. The policy commands the actuated hand joints from proprioceptive observations to maintain the cube in hand while following target orientations, testing whether Handroid can generate coordinated multi-finger contact behaviors beyond static grasping.

\begin{figure}[t]
    \centering
    \includegraphics[width=\textwidth]{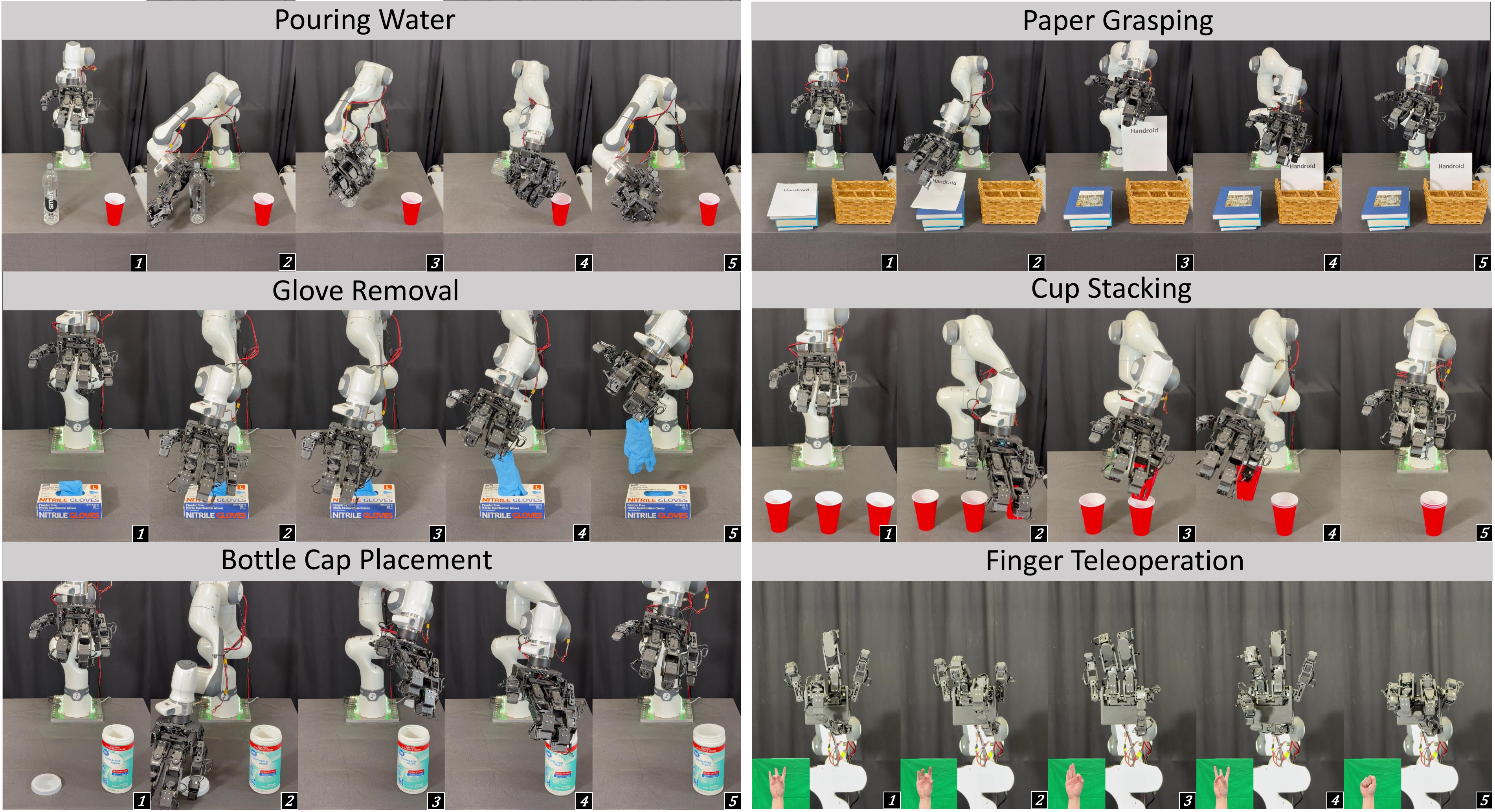}
    \vspace{-10pt}
    \caption{\textbf{Teleoperation as Dexterous Hand} of Handroid on various dexterous tasks.}
    \vspace{-8pt}
    \label{fig:Teleop Hand Task}
\end{figure}

\subsection{Humanoid}
\noindent\textbf{RL tracking control.}
We formulate humanoid locomotion as reference-guided tracking using time-indexed motions from two complementary sources. For planner-generated references, user-specified gait parameters define the footstep sequence, gait timing, and swing-foot height. The ZMP planner alternates parameterized single- and double-support phases, constructs a support-consistent desired ZMP sequence from the planned footstep locations, and generates the corresponding center-of-mass (CoM) trajectory using a fixed-height linear inverted pendulum model (LIPM) with LQR-based preview control~\cite{kajita2003biped}. Smooth swing-foot trajectories are synchronized with the same contact schedule. At each reference frame, Mink numerically solves the inverse-kinematics (IK) problem defined by the planned CoM position together with the torso and foot-pose targets, yielding the corresponding leg joint angles~\cite{zakka2026mink}. The resulting joint configurations and associated body states form a time-indexed reference motion. Alternatively, temporally interpolated motions authored with the Keyframe Editor provide a second source of tracking references~\cite{yang2026locomotion}. Given either type of reference, we train a closed-loop policy in MuJoCo that combines the current reference state with proprioceptive observations and outputs joint-position targets under simulated contact dynamics and actuator constraints~\cite{peng2018deepmimic}. The tracking objective rewards agreement between the reference and simulated root and body states:
\begin{equation}
% \small
r_t^{\mathrm{track}} =
0.5 r_t^{\mathrm{root,pos}}
+0.5 r_t^{\mathrm{root,ori}}
+r_t^{\mathrm{body,pos}}
+r_t^{\mathrm{body,ori}}
+r_t^{\mathrm{body,lin}}
+r_t^{\mathrm{body,ang}} .
\label{eq:track-reward}
\end{equation}

\noindent\textbf{RL velocity control.}
As a complementary reference-free approach, we train a velocity-conditioned policy without a predefined time-indexed motion trajectory~\cite{gu2024humanoidgym,he2025asap}. Conditioned on the commanded planar CoM velocity and yaw rate together with proprioceptive observations, the policy outputs joint-position targets. Unlike the tracking policy, it is optimized from command-conditioned rewards rather than motion imitation and does not use the ZMP planner, Mink, or an offline reference motion. Its objective combines planar-velocity and yaw-rate tracking with regularization:
\begin{equation}
% \small
r_t^{\mathrm{vel}} =
w_v
\exp\left(
-\frac{
\left\|
\mathbf{v}_{xy,t}
-
\mathbf{v}^{d}_{xy,t}
\right\|^2
}{\sigma_v^2}
\right)
+
w_\omega
\exp\left(
-\frac{
\left(
\omega_{z,t}
-
\omega^d_{z,t}
\right)^2
}{\sigma_\omega^2}
\right)
+
r_t^{\mathrm{reg}} .
\label{eq:velocity-reward}
\end{equation}

\noindent\textbf{Keyframe motion control.}
To support rapid whole-body motion authoring, we develop a Viser-based Keyframe Editor for Handroid~\cite{yang2026locomotion}. Users specify joint-space keyframes and their timing, from which the editor constructs dense joint-position trajectories through temporal interpolation. These trajectories can either be streamed through the robot's position-command interface for direct real-world execution or exported as user-authored references for RL tracking. The editor therefore provides a common motion-authoring interface for direct hardware execution and reference-based policy training.

\begin{figure}[t]
    \centering
    \includegraphics[width=\textwidth]{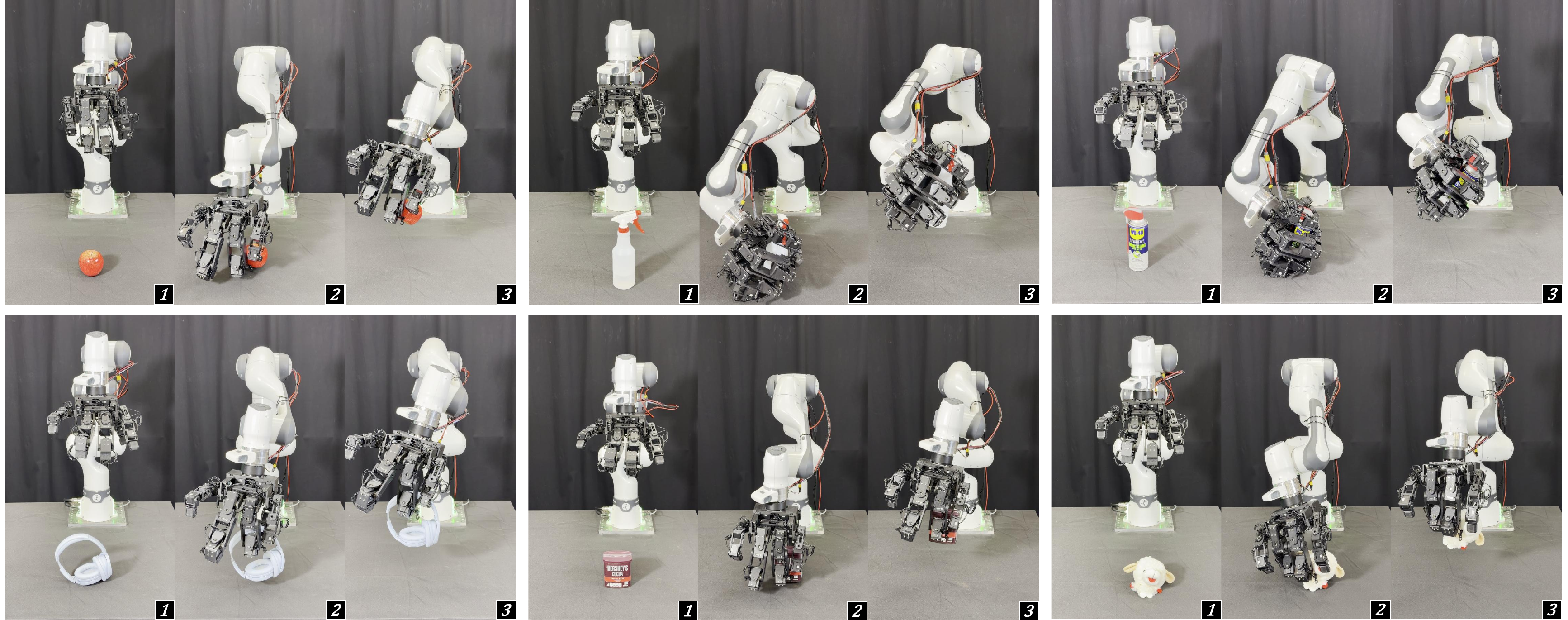}
    \vspace{-12pt}
    \caption{\textbf{Representative Diffusion Policy grasping rollouts}, conditioned on object point clouds and proprioception across randomized object poses.}
    \vspace{-12pt}
    \label{fig:DP Task Paper}
\end{figure}

\begin{table}[t]
    \centering
    \renewcommand\arraystretch{1.25}  % 行距
    \captionsetup{justification=centering, singlelinecheck=false}
    \caption{Real-world grasping results on 10 different objects.}
    \vspace{3pt}
    \resizebox{\linewidth}{!}{
        \begin{tabular}{cccccccccc}
            \toprule
            \textbf{Apple} & \textbf{Band Aid} & \textbf{Canister} & \textbf{Chip Tube} & \textbf{Cocoa Box} & \textbf{Earphone} & \textbf{Glove Box} & \textbf{Sheep} & \textbf{Sprayer} & \textbf{WD-40}
            \\
            8/10 & 6/10 & 8/10 & 9/10 & 7/10 & 6/10 & 7/10 & 9/10 & 5/10 & 7/10
            \\
            \bottomrule
        \end{tabular}
    }
    \label{tab:grasp}
    \vspace{-12pt}
\end{table}

\section{Experiments}

Our experiments are organized around three questions:
\begin{itemize}[leftmargin=*, itemsep=0.3em, topsep=0.3em]
    \item \textbf{Q1:} Can a reconfigurable body retain the fine-grained manipulation capabilities of a dedicated dexterous hand?
    \item \textbf{Q2:} Can the same body support stable locomotion and whole-body interaction when configured as a humanoid?
    \item \textbf{Q3:} Can embodiment reconfiguration from the same physical body extend the task space beyond what either embodiment can achieve alone?
\end{itemize}
\subsection{Dexterous Hand}
\noindent \textbf{Teleoperation}. We qualitatively evaluate the VR-based teleoperation interface through a set of real-world manipulation tasks, as shown in Fig.~\ref{fig:Teleop Hand Task} and Fig.~\ref{fig:teleop-hand-appendix}. The figure first illustrates the hand retargeting results, where Handroid follows different finger configurations of the operator in real time. We further demonstrate coordinated arm-hand teleoperation across tasks that require diverse manipulation properties, including precise grasping, object placement, stacking, hanging, articulated contact, deformable-object interaction, and liquid pouring. These demonstrations show that the interface can support both fine finger motion retargeting and whole arm-hand coordination, providing responsive and stable control for collecting demonstrations used in downstream policy learning.

\noindent \textbf{Dexterous Grasping}. We evaluate Handroid's grasping capability in a real-world setup with a Franka Research 3 robot arm. The task requires Handroid to grasp objects with diverse geometries in its dexterous hand embodiment. We select 10 objects with various shapes and sizes to evaluate the effectiveness of the learned policy.

We use scanned object meshes and RGB-D input from a RealSense L515 camera for FoundationPose~\cite{wen2024foundationpose} to estimate the 6D object pose. Given the estimated pose, the corresponding object mesh is transformed into the robot frame, and 512 surface points are sampled as the object point cloud input to the policy. This provides a consistent object-conditioned representation across training and evaluation.

We collect 10 demonstrations for each object using our VR-based teleoperation interface, resulting in 100 demonstrations in total. The policy is trained on these demonstrations with object point clouds as conditioning input. During evaluation, object poses are randomized within a preset workspace and differ from those in the demonstrations, testing whether the policy can generalize across object placement variations rather than simply replaying memorized trajectories.

Tab.~\ref{tab:grasp} reports the real-world grasping success rates across the 10 objects. Representative grasping rollouts and more details are shown in Fig.~\ref{fig:DP Task Paper} and Fig.~\ref{fig:DP Task Appendix}. The policy achieves an average success rate of 72\%, showing that the learned object-conditioned policy can perform dexterous grasping across objects with varied geometries and poses.

\begin{figure}[t]
    \centering
    \includegraphics[width=\textwidth]{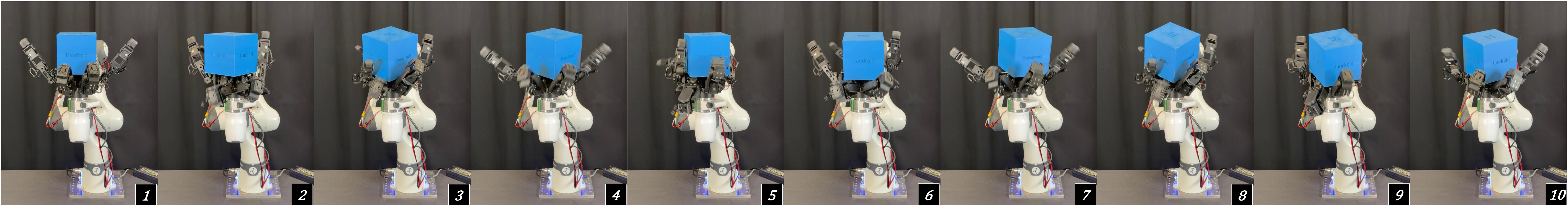}
    \vspace{-12pt}
    \caption{\textbf{Real-world cube reorientation} by Handroid in the Dexterous-Hand embodiment.}
    % \vspace{-12pt}
    \label{fig:Rotation Cube}
\end{figure}

\begin{figure}[t]
    \centering
    \includegraphics[width=\textwidth]{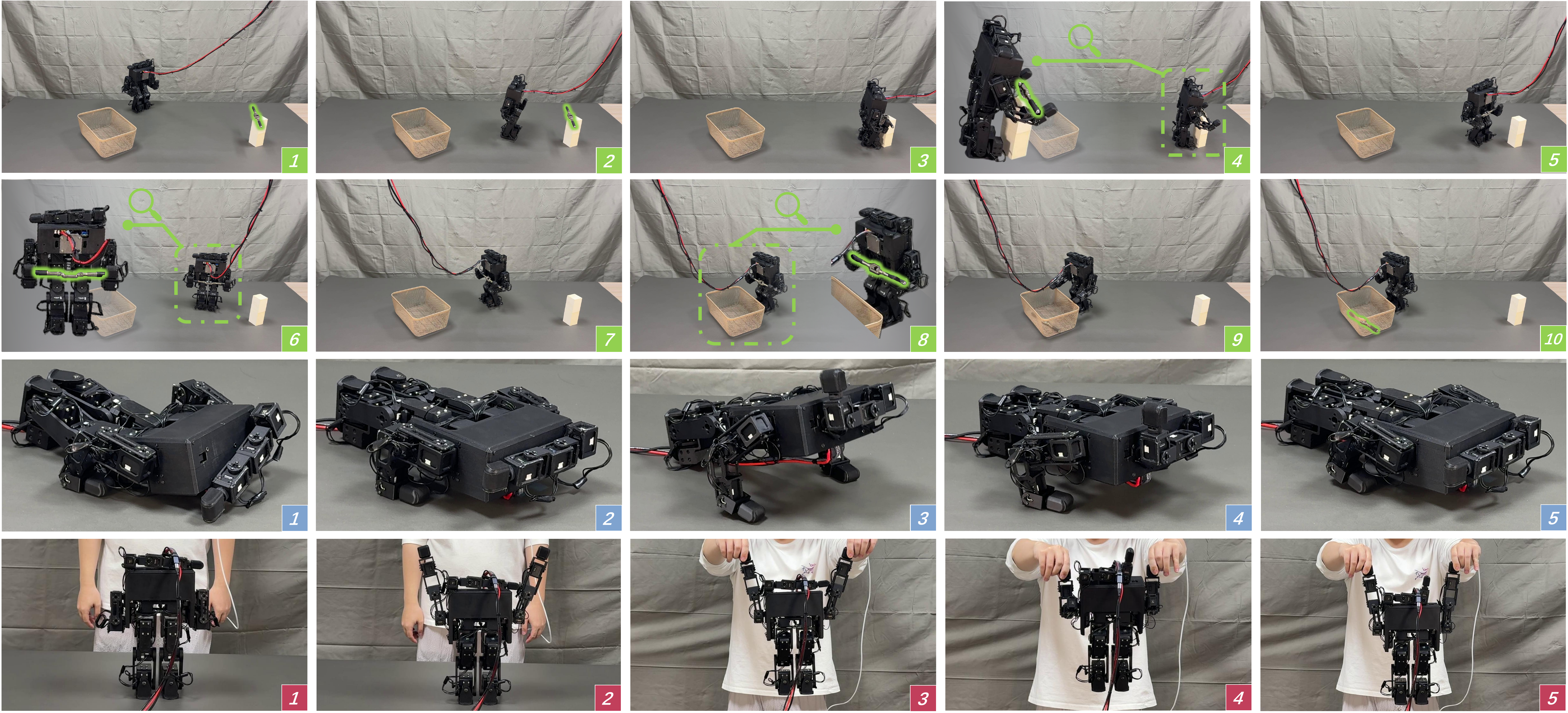}
    \vspace{-12pt}
    \caption{\textbf{Tasks in the Humanoid embodiment}, including pick-and-place, pull-up, and push-up.}
    \label{fig:Human Task}
    % \vspace{-12pt}
\end{figure}

\noindent \textbf{In-hand Reorientation}.
In addition to imitation learning, we also evaluate the in-hand reorientation policy on the real Handroid system. The Handroid in its hand embodiment is mounted to a Franka Research 3 arm, whose joints are positioned and fixed such that Handroid's palm faces upward. We 3D print a cube with the same dimensions as the object used during simulation training. At the beginning of each rollout, the Handroid motors are initialized to the same nominal joint configuration used in simulation, and the cube is placed manually in the hand without requiring a precise object pose. The policy receives the same proprioceptive observation history as in simulation and runs at 30 Hz on the real robot. We qualitatively evaluate whether the policy can maintain the cube in-hand. Representative real-world rollouts are shown in Fig.~\ref{fig:Rotation Cube}.

\noindent \textbf{Simulated RL Tracking.}
We evaluate the reference-tracking policy in MuJoCo using the ZMP-generated walking motion described in the Control section. During evaluation, the simulated robot states are compared with the corresponding time-aligned reference states. The learned policy achieves a joint-position tracking error of $0.12~\mathrm{rad}$ and a body-position tracking error of $0.0019~\mathrm{m}$. These results indicate close joint- and body-level agreement with the evaluated reference motion under simulated contact and actuator dynamics.

\noindent\textbf{Simulated RL Velocity Control.}
To complement reference-guided tracking, we further train and evaluate a reference-free velocity policy in MuJoCo. Rather than following a predefined time-indexed motion, the policy is conditioned on a commanded forward CoM velocity and proprioceptive observations. It outputs joint-position targets, which are applied through the same simulated actuation interface used by the tracking policy. Under a commanded forward velocity of $0.20~\mathrm{m/s}$, the policy achieves a velocity-tracking error of $0.052~\mathrm{m/s}$ with respect to the realized forward velocity. Qualitatively, the learned gait exhibits relatively short steps and a high stepping frequency. These results demonstrate velocity-conditioned locomotion under the evaluated command without relying on a predefined reference motion.
\begin{figure}[t]
    \centering
    \vspace{-10pt}
    \includegraphics[width=\textwidth]{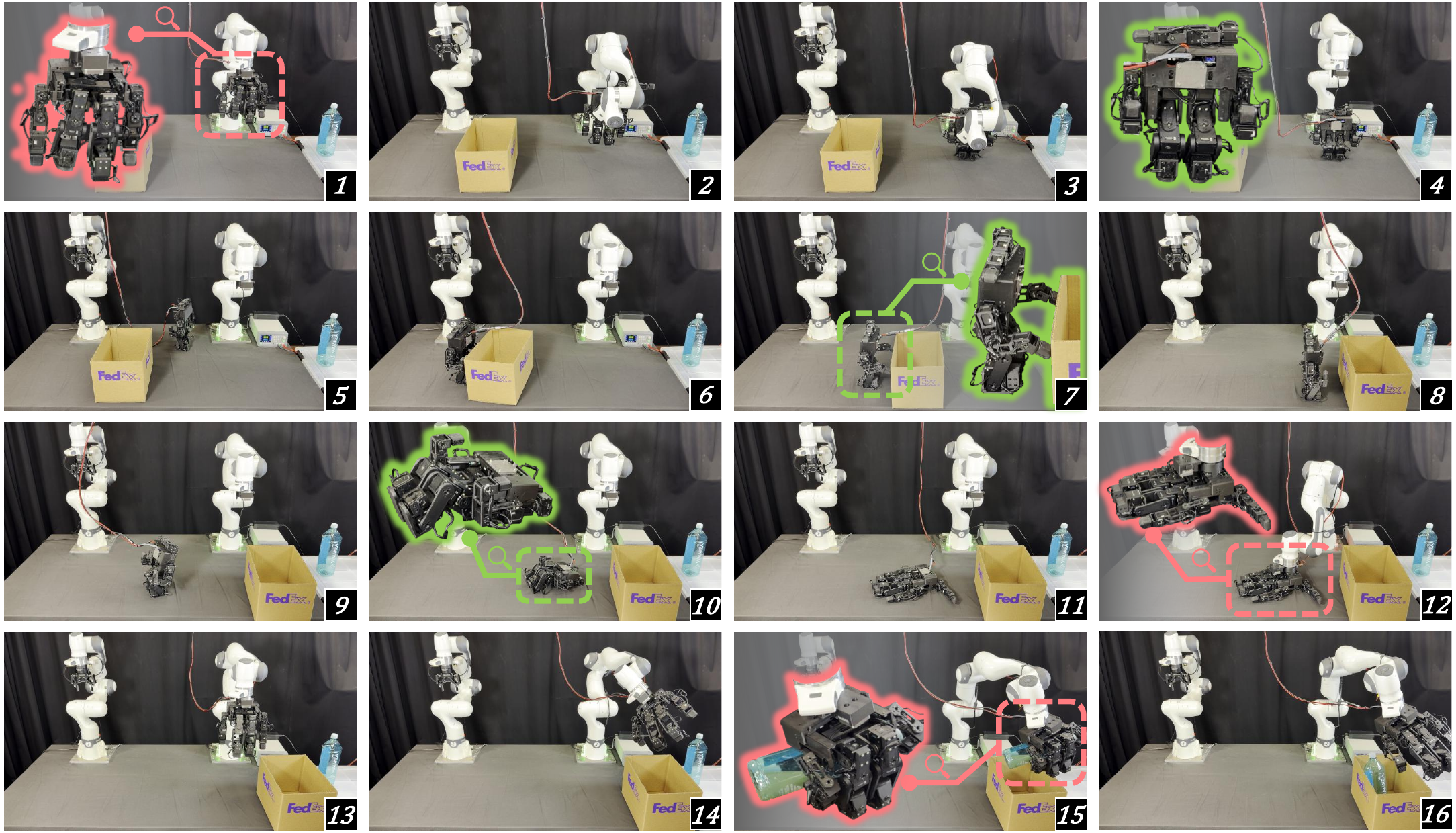}
    \vspace{-12pt}
    \caption{\textbf{Long-horizon task with Handroid}, including embodiment switching, detachment from the Franka, obstacle-avoidance walking, box pushing, lying down, re-docking with the Franka end effector, object grasping, and pick-and-place in the Dexterous-Hand embodiment.}
    \label{fig:long_horizon_task}
    \vspace{-1.5em}
\end{figure}

\vspace{3pt}
\noindent \textbf{Real-World Keyframe Motions.}
We qualitatively evaluate keyframe-authored motions on the physical Handroid in its Humanoid embodiment. Prerecorded motion clips are selected and triggered using either a gamepad or a keyboard. For walking, we construct a cyclic trajectory from six manually authored keyframes and set the interval between adjacent keyframes to $0.045~\mathrm{s}$ using the Temporal Interval Manager. The interpolated joint-position targets are first previewed in simulation and then streamed directly to the robot through the Real-Robot Execution Interface. As shown in Fig.~\ref{fig:Human Task}, Handroid executes push-ups, pull-ups, and a pick-and-place task in which it places a metal handle into a target box. We additionally demonstrate forward and backward walking, turning, and sidestepping. These demonstrations evaluate the Keyframe Editor as a real-robot motion-authoring and execution interface; the motions use direct joint-position playback and do not invoke either learned policy. Visualizations of all simulation experiments in the Humanoid embodiment can be found in Fig.~\ref{fig:Sim}.

\subsection{Dual Embodiment}
In this experiment, we evaluate the integrated execution capability of Handroid through a long-horizon task. The full execution process is shown in Fig.~\ref{fig:long_horizon_task}. First, in panels 1 to 4, Handroid switches from the Dexterous-Hand embodiment to the Humanoid embodiment and detaches from the Franka arm through the electromagnetic flange. Then, in panels 4 to 8, Handroid moves forward and turns in the Humanoid embodiment, bypasses the obstacle, reaches the back of the box, and pushes the box into the Franka workspace. Next, in panels 9 to 12, Handroid switches back from the Humanoid embodiment to the Dexterous-Hand embodiment and re-docks with the Franka arm through the electromagnetic flange. Finally, in panels 13 to 16, Handroid grasps the bottle and places it into the box in the Dexterous-Hand embodiment. The complete process demonstrates Handroid's ability to coordinate embodiment switching, humanoid locomotion, external-arm docking, and dexterous manipulation within a unified workflow. The electromagnetic flange design is shown in Fig.~\ref{fig:Flange}.

\section{Conclusion}
We present Handroid, a 27-DoF desktop-scale dual-embodiment robot that reuses a single electromechanical system as both a Dexterous Hand and a Humanoid. In the Dexterous-Hand embodiment, Handroid supports VR teleoperation, object-conditioned imitation learning from 100 demonstrations, real-world grasping with a 72\% average success rate, and in-hand cube reorientation. In the Humanoid embodiment, it supports ZMP-guided RL tracking control, RL velocity control and keyframe motion control. A long-horizon task further demonstrates bidirectional embodiment switching, humanoid object interaction, electromagnetic detachment and re-docking with a Franka Research 3 arm, and dexterous pick-and-place. These results establish Handroid as a compact platform for studying shared morphology, control, and learning across dexterous manipulation and humanoid motion.

\textbf{Limitation and Future Work.}
Handroid is an initial step toward dual-embodiment robot learning. Future wireless operation will reduce cable-induced disturbances and enable longer mobile tasks. Further miniaturization of actuators and structures can improve the Dexterous-Hand embodiment while preserving humanoid functionality. Adding fingertip cameras and tactile sensors will benefit both contact-rich manipulation and foot-ground contact estimation, since the same modules serve as fingers and feet. Handroid opens opportunities for cross-embodiment policy transfer, shared sensing, and unified learning frameworks that connect hand-scale dexterity with body-scale mobility.

% \section{Introduction}
	
%     Submission to CoRL 2026 will be entirely electronic, via a web site (not email). Information about the submission process and \LaTeX{} templates are available on the conference web site at \url{https://corl.org/}. For camera ready submission, use the \texttt{final} option for the \texttt{\textbackslash usepackage} command. 

%===============================================================================

% \clearpage
% % The acknowledgments are automatically included only in the final and preprint versions of the paper.
% \acknowledgments{If a paper is accepted, the final camera-ready version will (and probably should) include acknowledgments. All acknowledgments go at the end of the paper, including thanks to reviewers who gave useful comments, to colleagues who contributed to the ideas, and to funding agencies and corporate sponsors that provided financial support.}

%===============================================================================

% \clearpage
\appendix
% \newpage
\section*{\LARGE Appendix}
\renewcommand{\thesection}{\Roman{section}}
\setcounter{figure}{0}
\renewcommand{\thefigure}{A\arabic{figure}}
\section{Mechanical Design Details}
\label{app:structure-design}
\begin{figure}[t]
    \centering
    \includegraphics[width=\textwidth]{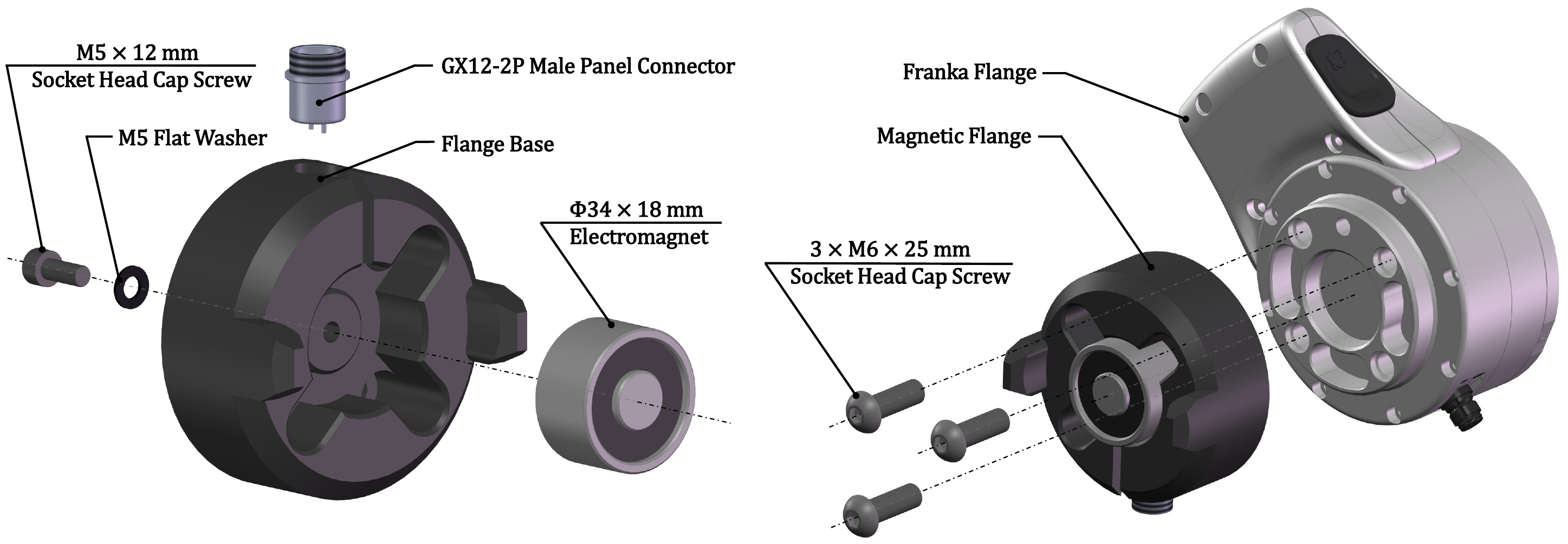}
    \caption{\textbf{Flange structure design.} Left: the magnetic flange mainly consists of an electromagnet and a flange base, where the electromagnet is fixed to the flange base by rear-side screws. Right: the connection method between the Franka flange and the magnetic flange.}
    \label{fig:Flange}
\end{figure}

\paragraph{Electromagnetic Flange.}
To enable rapid docking and detachment between Handroid and a Franka Research 3 arm, we design an electromagnetic flange, as shown in Fig.~\ref{fig:Flange}. The flange mainly consists of an electromagnet and a flange base, which are fastened to the Franka flange with screws. When energized, the electromagnet rapidly attaches to the magnetic plate mounted on Handroid. This connection provides approximately 180 N of holding force, which is sufficient to ensure stable attachment when Handroid operates in the Dexterous-Hand embodiment. Compared with conventional mechanical locking mechanisms, the electromagnetic flange has a simpler structure, faster connection, and lower alignment requirements, making it suitable for end-effector that require frequent docking and detachment.

% \renewcommand{\thesection}{\Roman{section}}
% \section{Actuator Selection}
\noindent \textbf{Actuator Selection.}
\label{app:actuation-selection}
Handroid uses three types of Dynamixel actuators: XC330-T288-T, XM430-W210-T, and 2XC430-W250-T. The compact XC330-T288-T actuators are mainly used in modules I, II, V, and VII, as well as in the transmission mechanism of module VI. The XM430-W210-T actuators provide higher peak torque and are placed in modules III and IV to support high-load joints such as hip pitch and ankle pitch. The 2XC430-W250-T actuators provide high torque with two orthogonal output axes. They are also used in modules III and IV to provide abduction/adduction and flexion/extension DoFs for the middle and ring fingers, and to provide hip-roll and knee-pitch DoFs in the Humanoid embodiment.

\renewcommand{\thesection}{\Roman{section}}
\section{Dexterous Hand Control Details}
\label{app:hand-control}
\paragraph{Teleoperation details.} The teleoperation system uses Apple Vision Pro hand tracking to obtain the operator's wrist pose and 22 hand keypoints. Finger commands are generated with the AnyTeleop retargeting framework~\cite{qin2023anyteleop} using a DexPilot-style objective~\cite{handa2020dexpilot}. The retargeting representation includes pairwise displacement vectors between fingertips and displacement vectors from each fingertip to the wrist, which provides a compact description of hand shape while reducing sensitivity to global hand motion. For the Franka arm, we record the initial operator wrist pose and the initial robot end-effector pose, and map the operator's relative wrist motion into the robot base frame to produce the target end-effector pose. Since the arm command only depends on relative pose mapping, the interface can be adapted to other robot arms with minimal changes. In our experiments, the combined arm-hand teleoperation loop runs stably above 20 Hz.

\begin{figure}[t]
    \centering
    \includegraphics[
        width=\linewidth,
        keepaspectratio
    ]{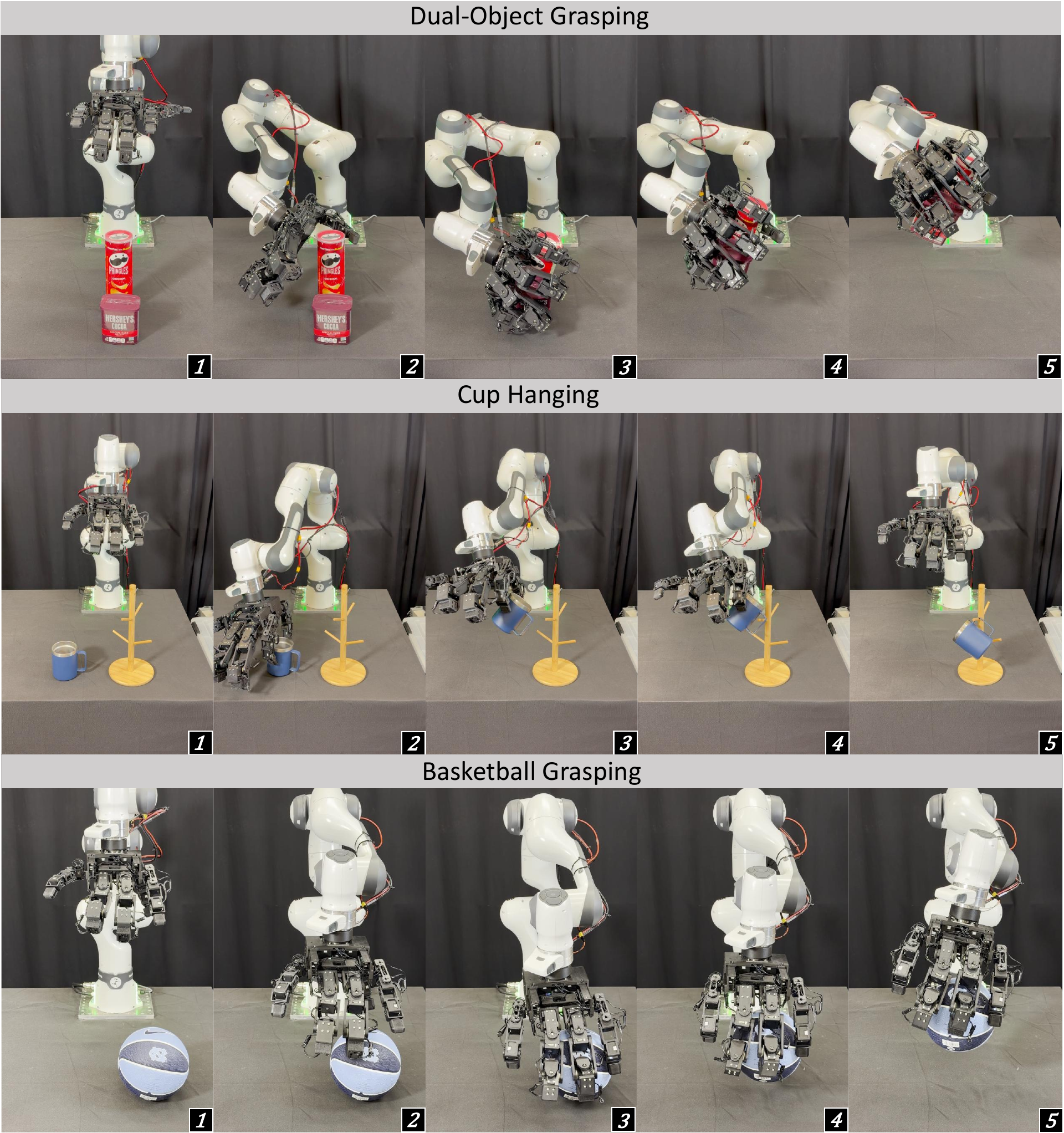}
    \caption{\textbf{Additional teleoperated dexterous-manipulation tasks.}
        From top to bottom: grasping and lifting two objects simultaneously in a
        single enveloping grasp; grasping and hanging a $270\,\mathrm{g}$ metal cup; and grasping a size-3 basketball.}
    \label{fig:teleop-hand-appendix}
    \vspace{-10pt}
\end{figure}
\begin{figure}[t]
    \centering
    \includegraphics[
        width=\linewidth,
        height=0.42\textheight,
        keepaspectratio
    ]{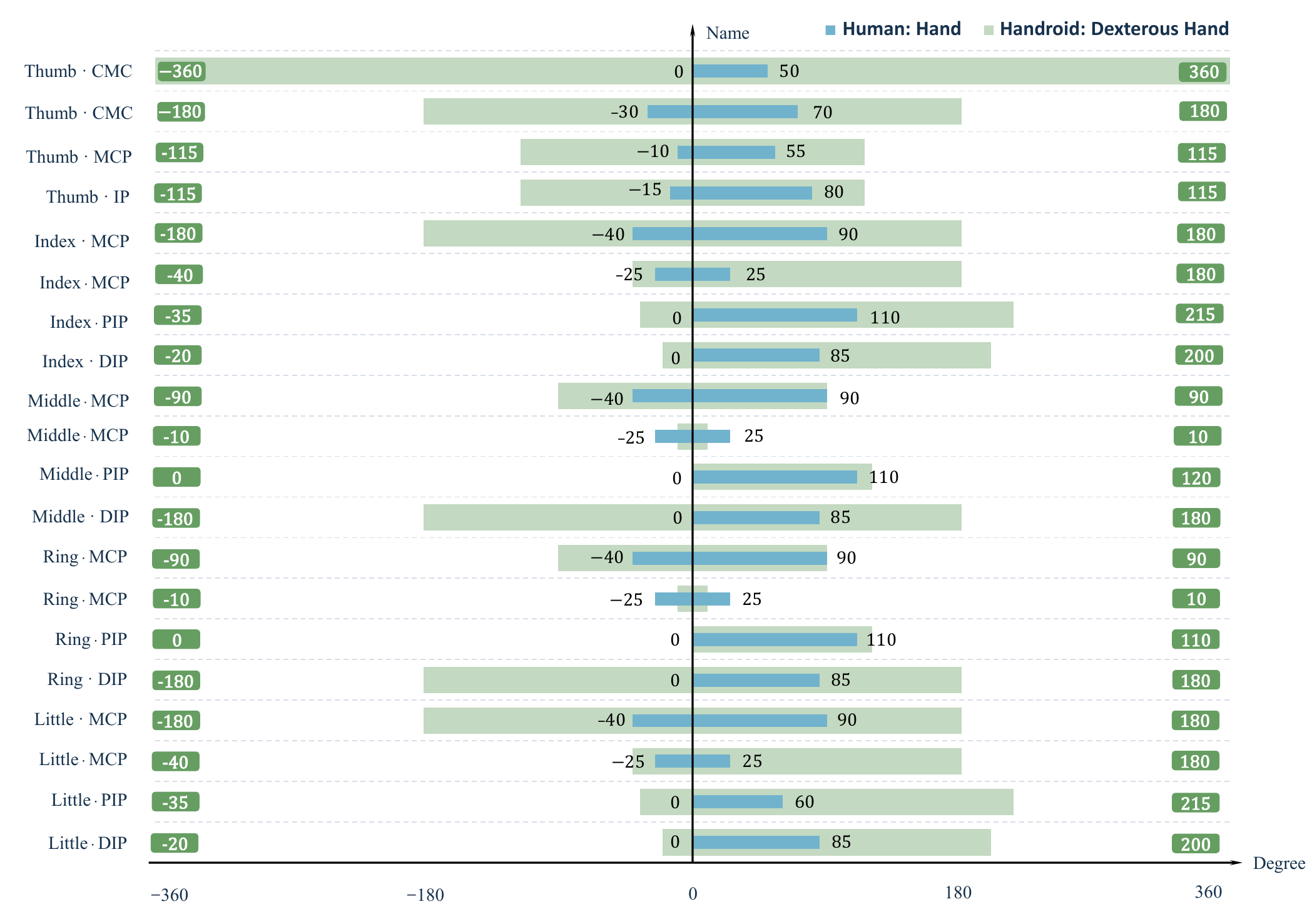}
    \caption{\textbf{RoM comparison: Human hand v.s.\ Handroid} with the Dexterous-Hand Embodiment. All available DoFs and their minimum and maximum joint angles are shown, demonstrating a DoF count comparable to the human hand and overall greater joint flexibility.}
    \label{fig:range-motion-hand}
    \vspace{-12pt}
\end{figure}

\begin{figure}[t]
    \centering
    \includegraphics[
        width=\linewidth,
        height=0.42\textheight,
        keepaspectratio
    ]{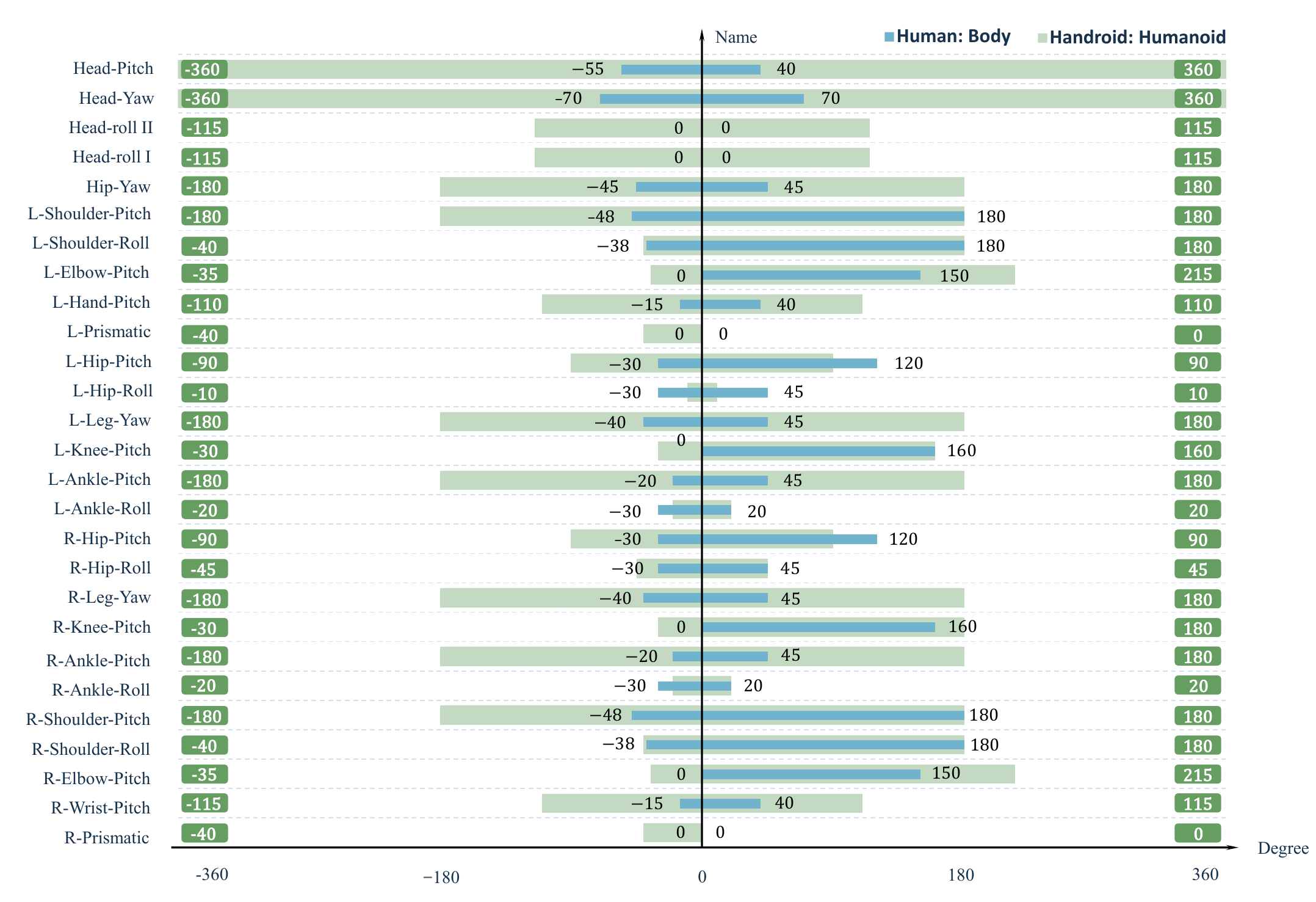}
    \caption{\textbf{RoM comparison: Human body v.s.\ Handroid with the Humanoid Embodiment.} Handroid exhibits broader RoM across most joint axes, indicating greater kinematic flexibility than the human body.}
    \label{fig:range-motion-humanoid}
    \vspace{-15pt}
\end{figure}
\vspace{-6pt}
\paragraph{Dexterous grasping policy.} Each grasping demonstration contains an object point cloud, robot proprioception, and an action trajectory for the arm-hand system. Given the object point cloud $\mathbf{P}^{O}$ and a proprioceptive history $\mathbf{s}_{t-n_{\mathrm{obs}}+1:t}$, the policy predicts an action chunk $\mathbf{a}_{t:t+n_{\mathrm{action}}-1}$. We encode the object point cloud with PointNet++~\cite{qi2017pointnet++} and encode the proprioceptive history with an MLP. The two feature vectors are concatenated and used as the conditioning input to a Diffusion Policy with a U-Net backbone~\cite{chi2025diffusion,ronneberger2015u}. In our implementation, the observation chunk size is $n_{\mathrm{obs}}=2$ and the action chunk size is $n_{\mathrm{action}}=8$. During deployment, overlapping predicted action chunks are combined with temporal ensembling to smooth the executed trajectory.

\begin{figure}[t]
    \centering
    \includegraphics[width=\textwidth]{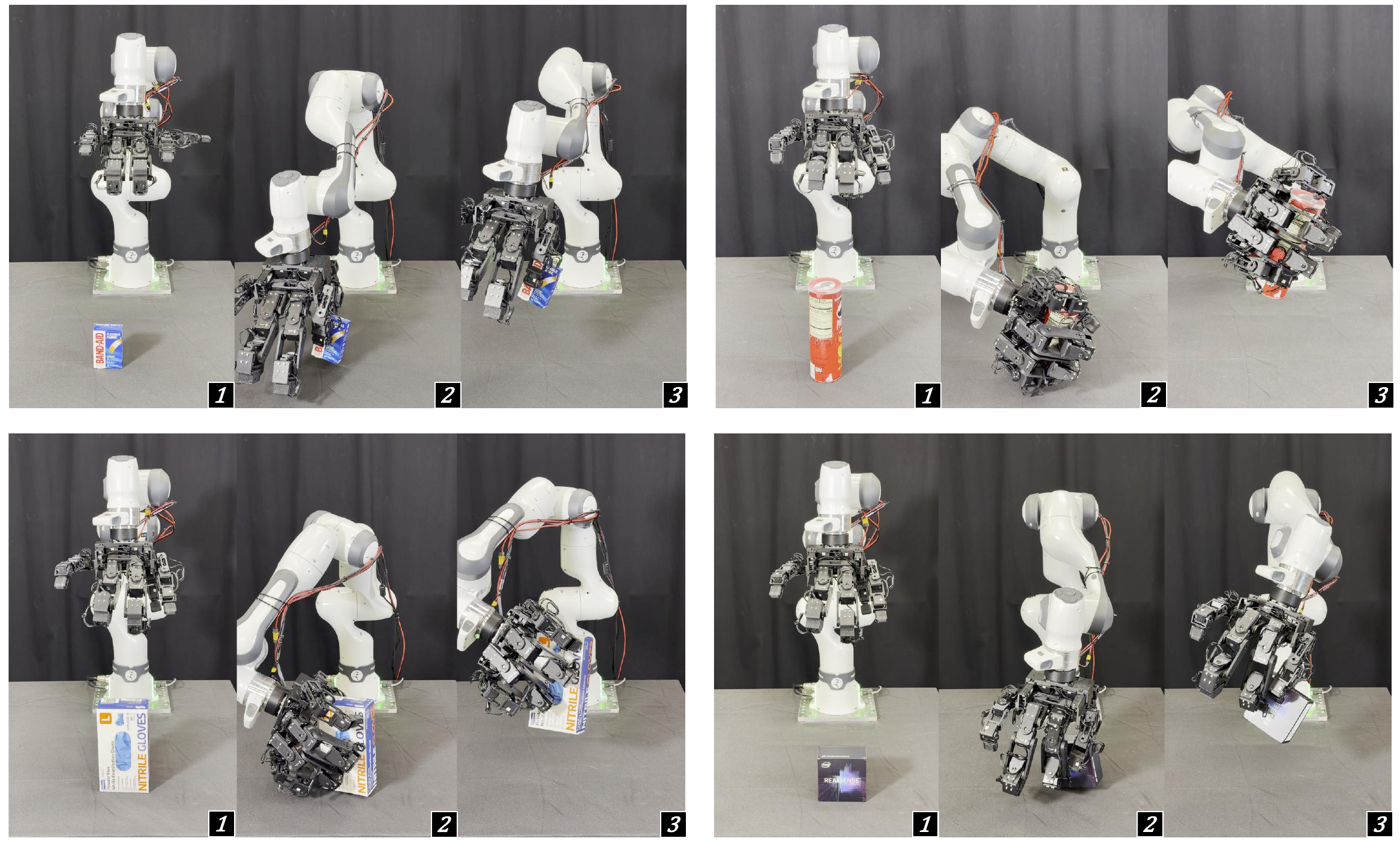}
    \caption{\textbf{Additional Diffusion Policy} grasping rollouts on the remaining four evaluated objects, including both fingertip picking and enveloping palm grasps of rigid and deformable objects.}
    \label{fig:DP Task Appendix}
    \vspace{-10pt}
\end{figure}
\vspace{-10pt}
\paragraph{In-hand reorientation policy.} The in-hand reorientation controller is trained in IsaacLab~\cite{mittal2025isaaclab} with PPO~\cite{schulman2017proximal}. The simulated setup uses a rooted, upward-facing Handroid hand and a cube initialized in a stable grasp. Joints that are not part of the hand task, including ankle, waist, and prismatic joints, are fixed. At each control step, the policy observes hand joint positions and recent command history, and outputs PD targets for the actuated hand DoFs. The target cube orientation is advanced by a fixed rotation after the current target is reached, forming a sequence of local reorientation goals. The reward encourages the cube to match a target orientation, so that continuous rotation emerges from a sequence of incremental reorientation goals. We additionally regularize action magnitude, unstable object motion, and deviation from the palm; a fall penalty discourages dropping the cube. For sim-to-real transfer, we randomize contact friction, actuator parameters, object mass and scale, and initial object and hand states during training.

\renewcommand{\thesection}{\Roman{section}}
\section{Humanoid Control Details}
\label{app:humanoid-control}

The tracking and velocity policies share the same position-target interface. For each controlled lower-body joint, the commanded position is computed as
\begin{equation}
q^{\mathrm{cmd}}_{j,t}
=
q^{\mathrm{default}}_{j}
+
s_j a_{j,t},
\label{eq:humanoid-action}
\end{equation}
where $j$ indexes the controlled lower-body joints and $t$ denotes the control step. Here, $q^{\mathrm{cmd}}_{j,t}$ is the commanded joint position, $q^{\mathrm{default}}_{j}$ is the default joint position, $a_{j,t}$ is the dimensionless policy action, and $s_j$ is the corresponding action scale. We set $s_j=0.25$ for all controlled lower-body joints; because revolute-joint positions are represented in radians, $s_j a_{j,t}$ corresponds to an angular offset from the default joint position. The simulated Dynamixel actuator model then converts the resulting position error and measured joint velocity into a PD torque subject to a velocity-dependent torque limit.
\vspace{-10pt}
\paragraph{RL tracking control.}
For planner-generated references, the gait schedule alternates between double- and single-support phases. Given a gait-cycle duration $T_{\mathrm{cyc}}$ and a prescribed single-to-double-support duration ratio $\rho$, the phase durations are
$T_{\mathrm{ds}}=T_{\mathrm{cyc}}/[2(\rho+1)]$ and
$T_{\mathrm{ss}}=\rho T_{\mathrm{ds}}$, respectively. Desired ZMP waypoints are constructed from successive planned footstep locations and synchronized with this contact schedule. Assuming a constant CoM height $h_{\mathrm{com}}$, the horizontal dynamics follow the fixed-height linear inverted pendulum model
\begin{equation}
\vspace{-2pt}
\mathbf{p}_{\mathrm{zmp}}
=
\mathbf{c}_{xy}
-
\frac{h_{\mathrm{com}}}{g}\ddot{\mathbf{c}}_{xy},
\label{eq:zmp-lipm}
\end{equation}
where $\mathbf{c}_{xy}$ and $\mathbf{p}_{\mathrm{zmp}}$ denote the planar CoM and ZMP positions, respectively, and $g$ denotes gravitational acceleration. An LQR-based preview controller balances desired-ZMP tracking, weighted by $Q_y$, against CoM-acceleration effort, weighted by $R$. During single support, the stance foot remains fixed while the swing-foot position and orientation follow smooth trajectories with a user-specified maximum height; during double support, both feet remain in contact. The foot and CoM trajectories follow the same phase schedule to maintain contact consistency during lift-off and touchdown.

For each planned frame, Mink solves a weighted differential-IK problem comprising CoM~\cite{zakka2026mink}, bilateral foot-pose, torso-stabilization, and posture-regularization tasks. The stance-foot task uses higher position and orientation weights than the swing-foot task to suppress support-foot motion, while non-leg joints are regularized toward their default configurations. Each frame is warm-started from the preceding solution and refined through 12 differential-IK iterations. The resulting leg joint angles and associated kinematic states form the planner-generated tracking reference.

Each reference motion is stored as a time-indexed sequence containing joint positions and velocities, body positions and orientations, and body linear and angular velocities. Handroid uses the hip link as the motion anchor; accordingly, the root terms in Eq.~\ref{eq:track-reward} are evaluated at this link. Planner-generated and keyframe-authored motions share this representation and are therefore processed by the same tracking pipeline.

At each control step, the tracking actor receives the current reference joint positions and velocities, the measured base angular velocity and projected gravity, and five-step histories of joint positions, joint velocities, and previous actions. The critic additionally receives reference-to-robot anchor offsets, robot body poses expressed in the anchor frame, and simulated base linear and angular velocities. The controlled lower-body joints comprise the hip-yaw joint and the bilateral hip-pitch, knee-pitch, ankle-pitch, and ankle-roll joints.

Each term in Eq.~\ref{eq:track-reward} applies an exponential kernel to its corresponding squared tracking error. The kernel widths for root position, root orientation, body position, body orientation, body linear velocity, and body angular velocity are $0.03~\mathrm{m}$, $0.15~\mathrm{rad}$, $0.03~\mathrm{m}$, $0.15~\mathrm{rad}$, $0.25~\mathrm{m/s}$, and $0.60~\mathrm{rad/s}$, respectively. The full objective additionally penalizes rapid action variation, joint-limit violations, and self-collisions. A rollout terminates when the reference motion ends, when the vertical hip-tracking error exceeds $0.04~\mathrm{m}$, when the absolute difference between the reference and simulated vertical components of the hip-frame projected-gravity vector exceeds $0.5$, or when the vertical tracking error of a monitored hand or foot exceeds $0.08~\mathrm{m}$.
\vspace{-6pt}
\begin{figure}[t]
    \centering
    \includegraphics[width=\textwidth]{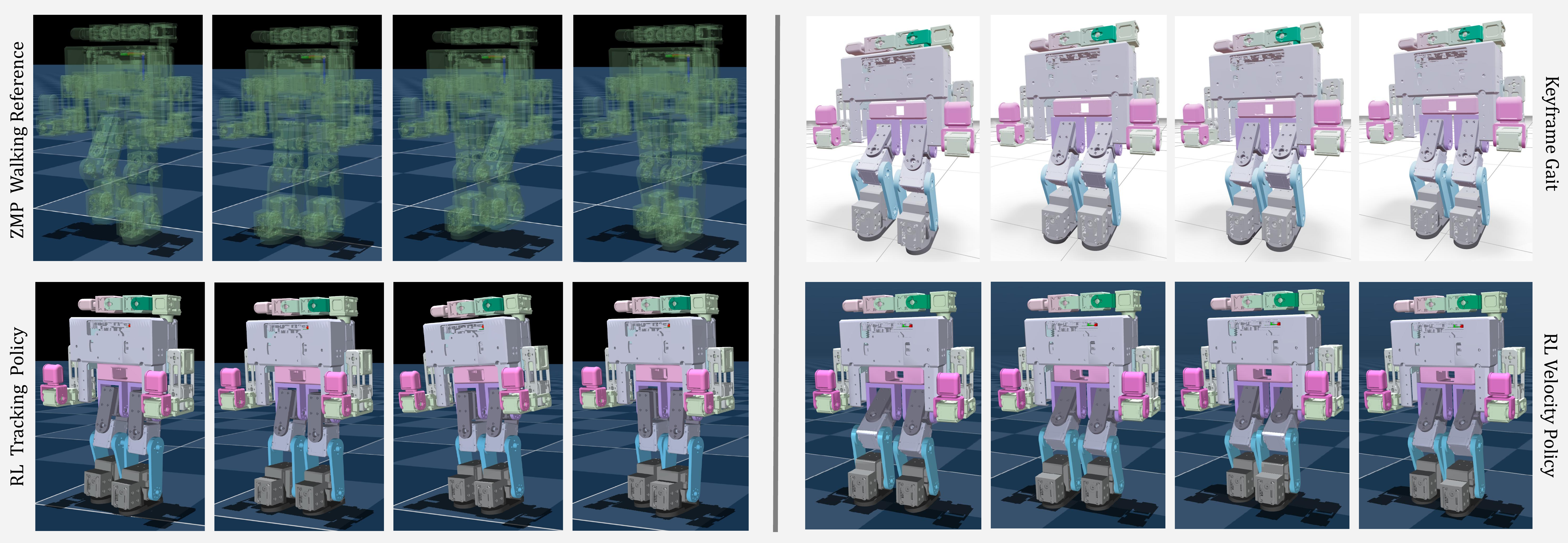}
    \vspace{-10pt}
    \caption{Simulation results of Humanoid locomotion, showing ZMP gaits, RL tracking gaits, RL from scratch gaits, and keyframe-generated gaits.}
    \vspace{-10pt}
    \label{fig:Sim}
    \vspace{-8pt}
\end{figure}
\vspace{-2pt}
\paragraph{RL velocity control.}
Unlike the tracking policy, the reference-free velocity policy is conditioned directly on locomotion commands rather than on a time-indexed motion reference, and it does not invoke the ZMP planner or Mink during training or inference. The actor observes the commanded planar CoM velocity and yaw rate, joint positions relative to their default values, joint velocities, and the previous action. During training, the critic additionally receives simulation-only privileged observations, including base linear and angular velocities, projected gravity, foot heights and air times, contact states, and contact forces. This asymmetric actor-critic design provides richer state information for policy learning without increasing observation requirements of the deployed actor.

The velocity-tracking objective is defined in Eq.~\ref{eq:velocity-reward}. Here, $\mathbf{v}_{xy,t}$ and $\omega_{z,t}$ denote the measured planar CoM velocity and yaw rate, respectively, while $\mathbf{v}^{d}_{xy,t}$ and $\omega^{d}_{z,t}$ denote their commanded values. The term $r_t^{\mathrm{reg}}$ collects auxiliary reward and penalty terms for maintaining an upright body orientation and a nominal joint posture, suppressing excessive body angular velocity, avoiding joint-limit violations, smoothing actions, regulating foot air time and swing-foot height and clearance, reducing foot slip, and encouraging soft landing. The planar-velocity and yaw-rate tracking terms use weights $w_v=w_\omega=2$, with kernel widths $\sigma_v=0.16~\mathrm{m/s}$ and $\sigma_\omega=0.50~\mathrm{rad/s}$, respectively. An episode terminates when it reaches the time limit or when the robot's tilt from the upright configuration exceeds $70^\circ$.
\vspace{-10pt}
\paragraph{Keyframe motion control.}
Each user-authored motion is represented by a sequence of joint-space keyframes and associated timestamps. Piecewise-linear interpolation produces a dense trajectory at the desired execution rate, and repeated segments can be packaged as cyclic action chunks. The Temporal Interval Manager modifies the timestamps without changing the key poses, allowing the motion rhythm to be adjusted interactively.

The editor supports two execution paths. For direct hardware execution, the interpolated joint-position targets are streamed through the Dynamixel position-command interface without invoking an RL policy. Alternatively, when a keyframe-authored motion is used for RL tracking, the dense trajectory is replayed in the robot model and converted through forward kinematics and finite differences into the joint and body states required by the common tracking-reference format. The same authored keyframes can therefore be used either for direct position playback on the physical robot or as reference motions for closed-loop tracking-policy training.

\bibliography{references}  % .bib

\end{document}